\documentclass[conference]{IEEEtran}
\IEEEoverridecommandlockouts
\usepackage{cite}
\usepackage{amsmath,amssymb,amsfonts}
\usepackage{algorithmic}
\usepackage{graphicx}
\usepackage{textcomp}
\usepackage{xcolor}
\def\BibTeX{{\rm B\kern-.05em{\sc i\kern-.025em b}\kern-.08em
    T\kern-.1667em\lower.7ex\hbox{E}\kern-.125emX}}
\begin{document}

\title{Synthetic Latent Fingerprint Generation Using Style Transfer\\
}

\author{\IEEEauthorblockN{Amol S. Joshi, Ali Dabouei, Nasser M. Nasrabadi, Jeremy Dawson}
\IEEEauthorblockA{\textit{Lane Department of Computer Science and Electrical Engineering} \\
\textit{West Virginia University}\\
Morgantown, USA \\
\{asj00003, ad0046\}@mix.wvu.edu, \{nasser.nasrabadi, jeremy.dawson\}@mail.wvu.edu}}

\maketitle

\begin{abstract}
Limited data availability is a challenging problem in the latent fingerprint domain. Synthetically generated fingerprints are vital for training data-hungry neural network-based algorithms. Conventional methods distort clean fingerprints to generate synthetic latent fingerprints. We propose a simple and effective approach using style transfer and image blending to synthesize realistic latent fingerprints. Our evaluation criteria and experiments demonstrate that the generated synthetic latent fingerprints preserve the identity information from the input contact-based fingerprints while possessing similar characteristics as real latent fingerprints. Additionally, we show that the generated fingerprints exhibit several qualities and styles, suggesting that the proposed method can generate multiple samples from a single fingerprint.  
\end{abstract}

\begin{IEEEkeywords}
Latent fingerprints, Synthetic latent fingerprint generation, Style transfer.
\end{IEEEkeywords}

\section{Introduction}
\begin{figure*}[htb]
\centering
    \includegraphics[width=0.59\linewidth]{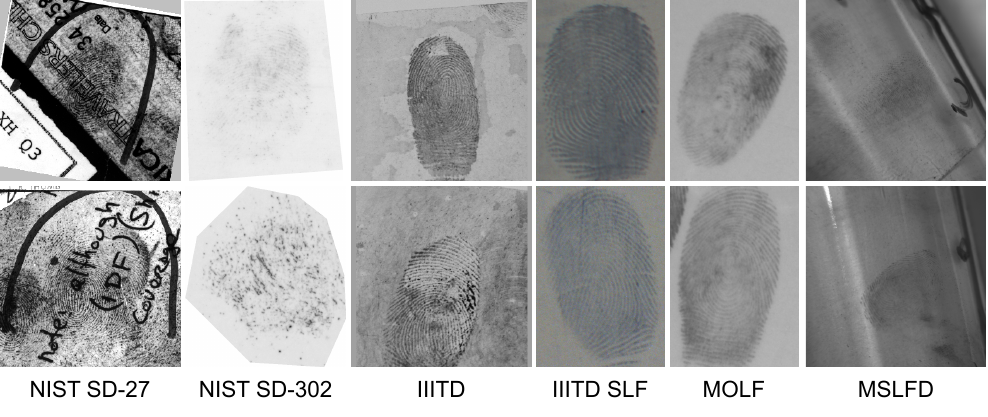}
    \caption{\label{fig:latent_samples}Latent fingerprints from datasets listed in Table \ref{tab1}.}
\end{figure*}

Fingerprints left on a surface unintentionally, also called latent fingerprints, play a vital role as evidence in forensic investigations. Unfortunately, these fingerprints are not readily viable for matching and recognition purposes. Due to the unconstrained environment at a crime scene and the complex acquisition process of latent fingerprints, they are notoriously indispensable to pre-processing, such as segmentation, enhancement, and feature extraction. Recent latent fingerprint pre-processing algorithms based on neural networks \cite{tang2017fingernet, nguyen2018robust, dabouei2018id, liu2020automatic, zhu2023fingergan} require larger datasets for training. However, the collection of latent fingerprints is an expensive and cumbersome task. Table \ref{tab1} summarizes the latent fingerprint datasets widely used for training and evaluating latent enhancement algorithms. The fingerprints in these datasets are deposited under controlled or uncontrolled conditions and lifted from various surfaces. NIST SD-302 dataset contains a large number of latent fingerprints, but not all of them have mated fingerprints. Moreover, the latent fingerprints in this dataset are substantially distinct from other datasets. Samples from these datasets are provided in Figure \ref{fig:latent_samples}. Combining these datasets for training pre-processing algorithms may introduce a class imbalance. Further, it is essential to use real latent fingerprints to evaluate these methods.
\begin{table}[t]
\centering
\caption{\label{tab1}Summary of released latent fingerprint datasets in the literature.}
\begin{tabular}{l l l l}
\hline
Dataset & \# of samples & \# of surfaces & Availability \\
\hline
NIST SD-27 \cite{156881} & 258 & N/A & No\\
NIST SD-302 \cite{249671} & 10000 & 30 & Yes\\
IIITD \cite{6117525} & 1016 & 2 & Yes\\
IIITD SLF \cite{6374604} & 720 & 1 & Yes\\
MOLF \cite{sankaran2015multisensor} & 4400 & 1 & Yes\\
MSLFD \cite{7358773} & 551 & 8 & Yes\\
\hline
\end{tabular}
\end{table}

This scarcity of data leads to the need for the generation of synthetic latent fingerprints that can be used to train the models so that real data can be utilized for a fair evaluation. With more synthetic images, these latent fingerprints need to possess certain characteristics. It is crucial to have identity features such as meaningful ridge structure, fingerprint shape, and minutiae points and noise features such as noisy backgrounds, surface, texture, etc. Many latent fingerprint pre-processing algorithms resort to a naive approach of blending a sensor-collected fingerprint with a noisy background to mimic a latent fingerprint \cite{dabouei2018id, liu2020automatic, huang2020latent}. Zhu et al. \cite{zhu2023fingergan} extend the weighted combination approach by applying plastic distortion \cite{Cappelli2001ModellingPD} on high-quality rolled fingerprints. This image-blending approach preserves the identity but fails to generate realistic latent fingerprints. 
Another method of generating synthetic fingerprints involves CycleGAN \cite{zhu2017unpaired}, which uses Generative Adversarial Networks(GAN) to transform images from one domain to another. Authors in \cite{ozturk2022minnet, wyzykowski2023synthetic} trained CycleGAN to transform slap/rolled fingerprints into latent fingerprints. However, these methods have limited style generation capacity. Wyzykowski and Jain \cite{wyzykowski2023synthetic} use multiple CycleGAN models to generate multiple styles. This might be inconvenient if latent fingerprints with more styles and qualities are required. Nonetheless, these approaches generate pairs of latent and sensor-collected fingerprints, which is ideal for training algorithms that use image-to-image translation for latent fingerprint pre-processing.  

Our goal is to generate multiple styles of latent fingerprints using a single model. When lifted from surfaces like paper, cardboard, ceramic tiles, etc., latent fingerprints exhibit different characteristics than those lifted from plastic and metallic surfaces. Additionally, the interaction between the surface and the subject causes uneven ridge densities and orientations. Therefore, the synthetic fingerprint generator must be trained to learn these variations for generating fingerprints of different styles. To this aim, we pose latent generation as a style transfer task from latent fingerprints to sensor-collected fingerprints. The primary task is to transform the ridge patterns in sensor-collected fingerprints into the ridge patterns in real latent fingerprints. This can be achieved by learning the distribution of latent fingerprints and fusing the distribution parameters with the source fingerprints. We use adaptive instance normalization \cite{huang2017arbitrary} to infuse the learned parameters of the latent fingerprint domain while reconstructing the fingerprints. Further, we blend these transformed ridge patterns with noisy backgrounds to manifest a similar distribution to the real latent fingerprints. For brevity, we will refer to sensor-collected fingerprints as fingerprints.

The generated latent fingerprints represent fingerprints lifted from different surfaces. Our contributions are three-fold:
\begin{itemize}
    \item We propose a simple and effective method that considers different surfaces and qualities while generating synthetic latent fingerprints.
    \item Our proposed method is flexible to generate multiple styles of the same fingerprint while preserving the underlying identity information.
    \item Our evaluation experiments demonstrate the similarities between synthetic and real latent fingerprints.
\end{itemize}

The paper is organized as follows; first, we discuss related work in Section \ref{sec:2}. The proposed method is described in Section \ref{sec:3} followed by a discussion on experiments and results in Section \ref{sec:4}. Finally, Section \ref{sec:5} concludes the paper.

\section{Related Work}\label{sec:2}
Many studies have been conducted to generate synthetic fingerprints. Before the advent of neural networks, hand-crafted feature-based approaches were developed to generate fingerprints. Capelli et al. \cite{1048096} used fingerprint shape, directional map, density map, and ridge patterns to create a master fingerprint. Further, they apply distortion, noise, and ridge variations to generate variants of the same master fingerprint. Zhao et al. \cite{6374554} proposed an approach based on statistics of fingerprint features such as type, size, ridge orientation, minutiae, and singular points. After generating a master print using the features, they apply non-linear plastic distortion and rigid transformations to get variants of the same fingerprint. Recent works typically use GANs to generate synthetic fingerprints \cite{bontrager2018deepmasterprints, minaee2018finger, bahmani2021high}. These methods focus on training GAN to learn the distribution of real fingerprints and generate synthetic fingerprints that contain the necessary identity information.

Style transfer is a way to learn to map the style of an image onto the contents of another image. Neural network-based style transfer is also explored in the image synthesis task \cite{men2020controllable, zhu2020sean, lyu2023dran}. Men et al. \cite{men2020controllable} developed a person image synthesis algorithm that encodes attributes such as pose, head, base, clothes, etc. The style code is then injected into the AdaIN \cite{huang2017arbitrary} features during decoding. Authors in \cite{zhu2020sean, lyu2023dran} proposed region adaptive normalization to control the style encoding in different image patches. This allows more flexibility to generate images with fine details. Despite these works, to the best of our knowledge, latent fingerprint synthesis has yet to be attempted with style transfer.

\section{Methodology}\label{sec:3}
A widely adopted conventional approach to generating synthetic latent fingerprints applies noise to good-quality fingerprints and blends them with noisy backgrounds. It uses the equation below:
\begin{equation}\label{eq1}
    \centering
    I_{latent} = \alpha \times I_{fingerprint} + (1 - \alpha) \times I_{noise}.
\end{equation}
However, in real-world scenarios, the latent fingerprints are lifted from multiple surfaces under unforeseeable environments. Depending on the nature of the surface and the action that caused the fingerprint to be left on the surface, the latent fingerprints exhibit different styles. As a result, using the blending method naively with good-quality fingerprints may not represent the distribution of real latent fingerprints. 

\begin{figure*}[t]
\centering
    \includegraphics[width=1\linewidth]{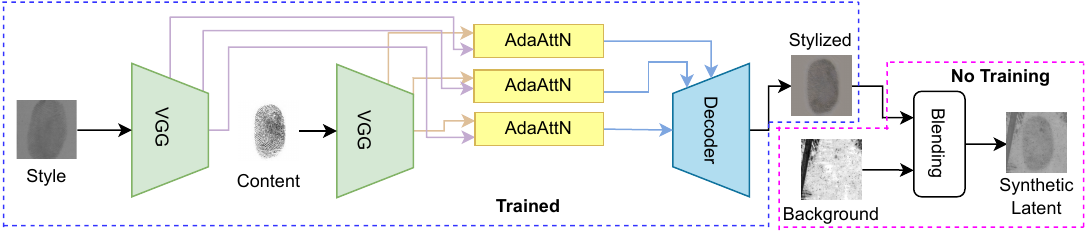}
    \caption{\label{fig:network}Architecture of the proposed method. The style transfer network is trained using real latent fingerprints and is marked by the blue box, whereas the image blending does not involve training and is represented in the magenta box.}
\end{figure*}

We propose learning the noise and distortions in ridge patterns acquired from multiple surfaces and transferring them to fingerprints to mimic the real latent fingerprints. To this aim, we devise a simple and efficient approach involving style transfer and image blending. Further, section \ref{sec3.1} illustrates the style transfer network, and section \ref{sec3.2} discusses image blending. Figure \ref{fig:network} illustrates the network architecture.

\subsection{Style Transfer}\label{sec3.1}
The style transfer module is responsible for extracting style from a latent fingerprint $F_s$ and fusing it with the content fingerprint $F_c$ during the reconstruction phase. We use AdaAttN \cite{liu2021adaattn} to learn the style of latent fingerprints and transform the fingerprint ridges to have a similar style. The style transfer network uses an encoder $E(.)$ to extract content and style embeddings. The extracted embeddings are then passed to the AdaAttN block, which adaptively transfers the style statistics to the content embeddings. The style transfer network uses several layers of pre-trained VGG-19 \cite{simonyan2014very} model to obtain embeddings with different spatial sizes and image characteristics. The AdaAttN block uses the attention mechanism to compute the weighted mean and variance map. Then, adaptive normalization proposed by Huang et al. \cite{huang2017arbitrary} is used to obtain the stylized features. Finally, the decoder reconstructs the stylized features to a synthetic latent fingerprint $F_{cs}$.  

The decoding at the end of the style transfer network may produce a random texture pattern to satisfy the objective function. Therefore, to preserve the identity information and the ridge structure of the fingerprints, we use another encoder $V(.)$ trained to extract features helpful in matching two fingerprints. The embeddings of the fingerprint $F_c$ and the stylized fingerprint $F_{cs}$ enforce the identity constraint during training. We trained the style transfer network using the following objective function:
\begin{equation}\label{eq2}
    \centering
    \mathcal{L} = \lambda_g\mathcal{L}_{gs} + \lambda_l\mathcal{L}_{lf} + \lambda_{i}\mathcal{L}_{id},
\end{equation}
where $\mathcal{L}_{gs}$ is a global style loss computed between the mean and standard deviation of embeddings of $F_s$ and $F_{cs}$ extracted using $E(.)$. $\mathcal{L}_{lf}$ is a local feature loss that minimizes the distance between features of $E(F_{cs})$, $E(F_c)$, and $E(F_s)$. Lastly, $\mathcal{L}_{id}$ is an identity constraint between $V(F_{cs})$ and $V(F_c)$. We use mean squared error to calculate the loss terms. We empirically set a value of $1.0$ for $\lambda_i$, whereas the default values of $3.0$ for $\lambda_g$ and $10.0$ for $\lambda_l$ are used during training. 

\subsection{Image Blending}\label{sec3.2}
The output of the style transfer network is distorted ridge patterns that appear similar to the ridge patterns in real latent fingerprints. However, we can profusely notice the noisy backgrounds and textured patterns in real latent fingerprints. Therefore, we incorporate the image blending from Eq. \ref{eq1} to generate realistic latent fingerprints. We replace $I_{fingerprint}$ by the output of the style transfer network and consider several background images cropped from real latent fingerprints as $I_{noise}$. During all the experiments, we set $\alpha$ between $0.3$ to $0.8$. This combination of style transfer network and image blending presents the flexibility to manipulate the style, quality, surface, and background of the generated fingerprints without retraining the network. Further, the synthetic fingerprints and the corresponding content fingerprints are spatially consistent. Therefore, the spatial features extracted from the fingerprint can be used as a target while training a neural network for latent fingerprint pre-processing. Later in section \ref{sec:4.3}, we discuss the effect of blending noisy background with the output of the style transfer network.

\section{Experiments}\label{sec:4}
In section \ref{sec:4.1}, we discuss the datasets used and generated for the evaluation experiments. Later, we describe the evaluation criteria and results in Section \ref{sec:4.1} and Section \ref{sec:4.2}, respectively.
\subsection{Datasets}\label{sec:4.1}
Training the style transfer network requires fingerprints as content images and latent fingerprints as style images. Therefore, we combined fingerprints from MOLF and MSLFD datasets totaling 12,444 for training and 600 for evaluation. For the style images, we used 4,400 latent fingerprints from MOLF, which has fingerprints lifted from ceramic tile \cite{sankaran2015multisensor}. Additionally, we included 170 latent fingerprints from two different surfaces from the MSLFD dataset. Further, we create pairs of latent fingerprints and content fingerprints such that they belong to the same finger of the same subject. This aids the identity preservation constraint in the objective function. We use latent fingerprints from IIITD and SLF datasets as the style images during evaluation experiments.

Once the style transfer network is trained, we generate 600 synthetic latent fingerprints. Finally, we create two sets, Synthetic-1 and Synthetic-2, using backgrounds from different surfaces and textures. The Synthetic-1 dataset represents latent fingerprints lifted from plain surfaces such as ceramic tiles and cardboard. In contrast, the Synthetic-2 dataset comprises latent fingerprints lifted from plastic and paper surfaces with printed text.
\subsection{Evaluation Criteria}\label{sec:4.2}
For evaluating a synthetic data generator, measuring the similarities between the synthetic and real data is imperative. We use various aspects of fingerprints for comparing the characteristics of real and generated latent fingerprints. First, we use quality distribution as a metric to demonstrate the similarity. To this aim, we use NFIQ 2.0 \cite{918416} to obtain the quality scores of latent fingerprints. The second metric is the similarity between the data distribution of real and synthetic fingerprints. We use t-Distributed Stochastic Neighbor embeddings (t-SNE) \cite{JMLR:v9:vandermaaten08a} to showcase the distribution of multiple datasets to compare with the synthetic fingerprints. t-SNE uses high-dimensional feature embeddings of size 512 and reduces the dimensionality to generate two components to visualize the distribution.

\begin{figure}[t]
\centering
    \includegraphics[width=6.5cm]{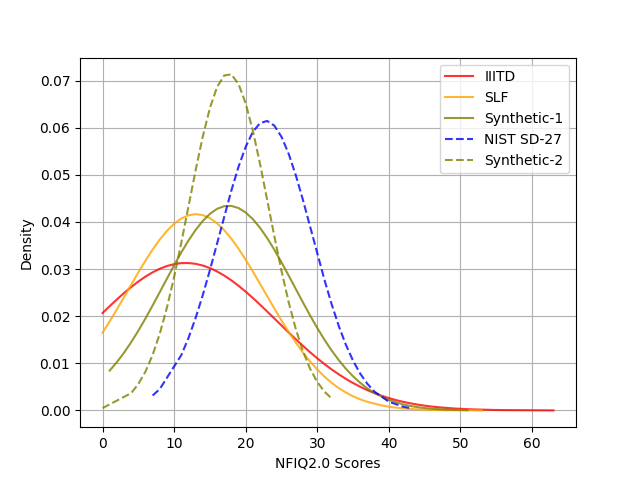}
    \caption{\label{fig:quality_dist}NFIQ 2.0 quality score distribution of multiple datasets. Solid and dashed lines represent datasets with different surfaces and styles.}
\end{figure}

Further, we study minutiae points to analyze the realistic nature of synthetic fingerprints. This analysis helps determine if the synthetic latent fingerprints have meaningful patterns and genuine minutiae points. We use the Verifinger SDK v10.0 \cite{verifinger} to extract minutiae and perform matching experiments. Lastly, we analyze the matching score distribution of genuine pairs consisting of synthetic latent and corresponding mated fingerprints. Due to the noisy and distorted nature of latent fingerprints, the recognition accuracy is relatively low compared to fingerprint matching. Comparing the matching scores of mated pairs helps estimate if the synthetic fingerprints are challenging enough for the matchers to extract features.

\subsection{Results}\label{sec:4.3}
\begin{figure}[t]
\centering
    \includegraphics[width=1\columnwidth]{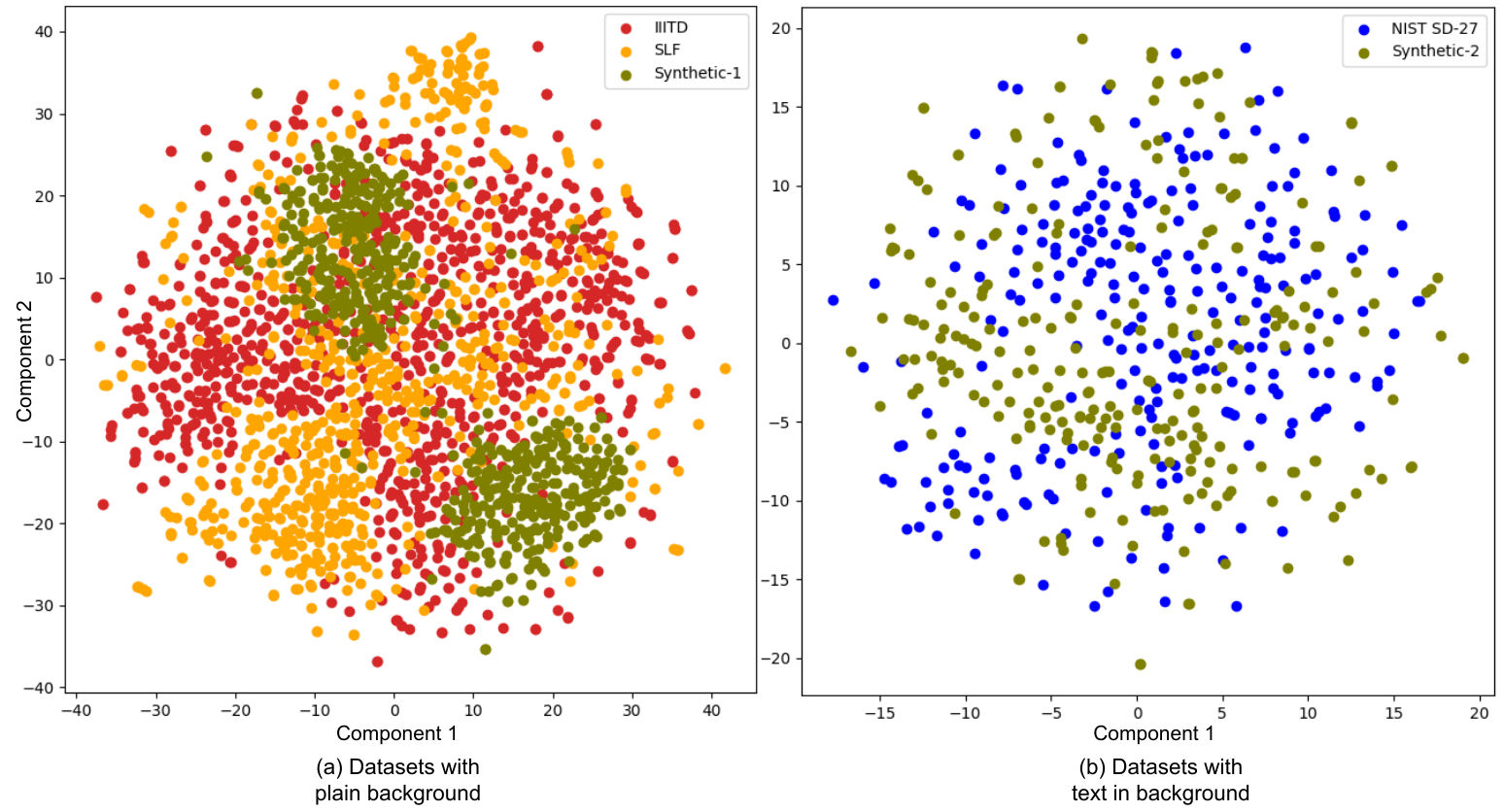}
    \caption{\label{fig:tsne}t-SNE distribution of multiple datasets. Plot (a) represents datasets with plain backgrounds from surfaces like ceramic tiles and cards. Plot (b) represents latent fingerprints lifted from plastic and paper with text in the background.}
\end{figure}

Determining the quality of latent fingerprints is crucial in matching and recognition scenarios. Due to the complex acquisition process of latent fingerprints, they often exhibit poor quality scores. In Figure \ref{fig:quality_dist}, we compare if the generated synthetic latent fingerprints have a similar quality score distribution with the real data. The latent fingerprints in IIITD and SLF datasets have a wide range of quality scores, whereas the NIST SD-27 dataset has a smaller range due to the arbitrary texture patterns and highly distorted ridge patterns. The plot suggests the closeness of quality levels among Synthetic-1, IIITD, and SLF datasets. Similarly, curves for NIST SD-27 and Synthetic-2 datasets also match each other. Next, we plot t-SNE to demonstrate the overlapping distribution of real and synthetic fingerprints. Figure \ref{fig:tsne} provides the distribution for multiple datasets of various styles. Note that in Figure \ref{fig:tsne}(a), the data points for the Synthetic-1 dataset are congregated in two regions. This behavior is due to the limited style references used to transform the ridge patterns during synthetic generation. At the same time, the arbitrary noise patterns in the real latent fingerprints make the distribution widespread. Regardless, both plots show evidence of the embeddings of the synthetic and real latent fingerprints in the high-dimensional space corresponding with datasets of respective styles. Further, this suggests that our proposed method can generate realistic latent fingerprints with real latent fingerprint characteristics.

\begin{figure*}[htb]
\centering
    \includegraphics[width=1\linewidth]{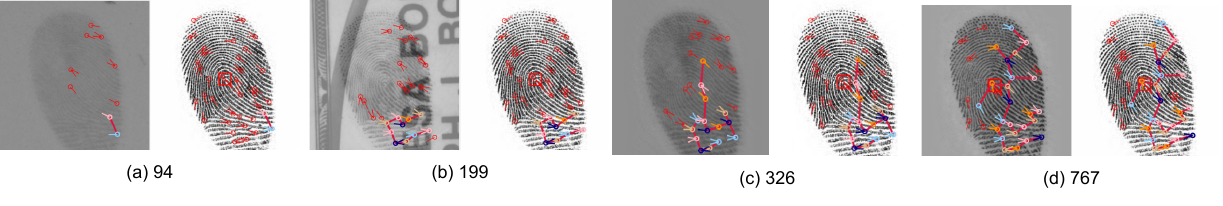}
    \caption{\label{fig:samples}Generated synthetic latent fingerprints of various styles and qualities. In each pair, the image on the left is the latent fingerprint, and on the right is the corresponding input fingerprint. The number mentioned at the bottom of each pair represents the matching score obtained using the Verifinger SDK v10.0. See Section \ref{sec:4.3} for additional details.}
\end{figure*}

Despite similar quality and t-SNE distributions, the synthetic latent fingerprints should represent some identity. Ideally, a synthetic latent fingerprint should have the same identity as the source fingerprint used as input to the style transfer network. Figure \ref{fig:samples} demonstrates the identity similarity between the synthetic latent and input fingerprint. It shows detected and correctly matched minutiae, suggesting that the proposed method preserves critical features such as the ridge structure and minutiae points. Further, the figure indicates the ability of the proposed method to generate multiple synthetic samples from the same fingerprint with varying quality and styles. 

To investigate the importance of the style transfer network and compare it with the naive approach of image blending used in \cite{dabouei2018id, liu2020automatic,huang2020latent}, we generated a set of synthetic latent fingerprints without using the style transfer network. We applied speckle noise to the fingerprints and blended them with noisy backgrounds. Then, we conducted a matching experiment with genuine pairs from this dataset. In Table \ref{tab2}, we compare the mean, standard deviation, and median of matching scores for genuine pairs of latent fingerprints generated by our method and the real latent fingerprint dataset. A significant difference between the distribution parameters shows that a weighted combination of a distorted fingerprint and noisy background is insufficient to model realistic latent fingerprints. The matcher can easily recognize the fingerprint despite the background noise.
\begin{table}[htb]
\centering
\caption{\label{tab2}The mean, standard deviation, and median of matching scores for genuine pairs belonging to different latent fingerprint datasets. VeriFinger SDK v10.0 was used to obtain the matching scores.}
\begin{tabular}{l l l l}
\hline
Latent dataset & Mean & Standard deviation & Median\\
\hline
W/o style transfer & 609.3629 & 381.131 & 572.0\\
Ours & 89.1046 & 101.1472 & 43.5\\
Real & 63.5454 & 47.9998 & 54.0\\
\hline
\end{tabular}
\end{table}

\section{Conclusion}\label{sec:5}
We proposed a simple and effective approach to synthetic latent fingerprint generation. We showed that the naive approximation of latent fingerprints inadequately represents real latent fingerprints. We revised it and proposed an algorithm to generate realistic latent fingerprints using a style transfer network to exploit the style features of real latent fingerprints and transform the ridge structure to appear as a latent fingerprint. Further, the stylized ridges are blended with noisy backgrounds for a better representation of real latent fingerprints. Our evaluation with various metrics suggests that the proposed method reliably generates latent fingerprints of various styles and qualities while preserving identity information. 

{\small
\bibliographystyle{ieeetr}
\bibliography{egbib}
}

\end{document}